# 14

# Visual Speech Recognition

Ahmad B. A. Hassanat
*IT Department, Mu'tah University,*
*Jordan*

## 1. Introduction

Lip reading is used to understand or interpret speech without hearing it, a technique especially mastered by people with hearing difficulties. The ability to lip read enables a person with a hearing impairment to communicate with others and to engage in social activities, which otherwise would be difficult. Recent advances in the fields of computer vision, pattern recognition, and signal processing has led to a growing interest in automating this challenging task of lip reading. Indeed, automating the human ability to lip read, a process referred to as visual speech recognition (VSR) (or sometimes speech reading), could open the door for other novel related applications.
VSR has received a great deal of attention in the last decade for its potential use in applications such as human-computer interaction (HCI), audio-visual speech recognition (AVSR), speaker recognition, talking heads, sign language recognition and video surveillance. Its main aim is to recognise spoken word(s) by using only the visual signal that is produced during speech. Hence, VSR deals with the visual domain of speech and involves image processing, artificial intelligence, object detection, pattern recognition, statistical modelling, etc.
There are two different main approaches to the VSR problem, the visemic* approach and the holistic approach, each with its own strengths and weaknesses. The traditional and most common approaches to automatic lip reading are based on visemes. A Viseme is the mouth shapes (or appearances) or sequences of mouth dynamics that are required to generate a phoneme in the visual domain. However, several problems arise while using visemes in visual speech recognition systems such as the low number of visemes (between 10 and 14) compared to phonemes (between 45 and 53). Visemes cover only a small subspace of the mouth motions represented in the visual domain, and many other problems. These problems contribute to the bad performance of the traditional approaches; hence, the visemic approach is something like digitising the signal of the spoken word, and digitising causes a loss of information.
The holistic approach such as the "visual words" (Hassanat, 2009) considers the signature of the whole word rather than only parts of it. This approach can provide a good alternative to the visemic approaches to automatic lip reading. The major problem that faces this approach is that for a complete English language lip reading system, we need to train the whole of the English language words in the dictionary! Or to train (at least) the distinct ones. This

* Related to a Viseme.

approach can be effective if it is trained on a specific domain of words, e.g. numbers, postcodes, cities, etc.
A typical VSR system consists of three major stages: detecting/localizing human faces, lips localization and lip reading. The accuracy of a VSR system is heavily dependent on accurate lip localisation as well as the robustness of the extracted features. The lips and the mouth region of a face reveal most of the relevant visual speech information for a VSR system (see Figure 1).

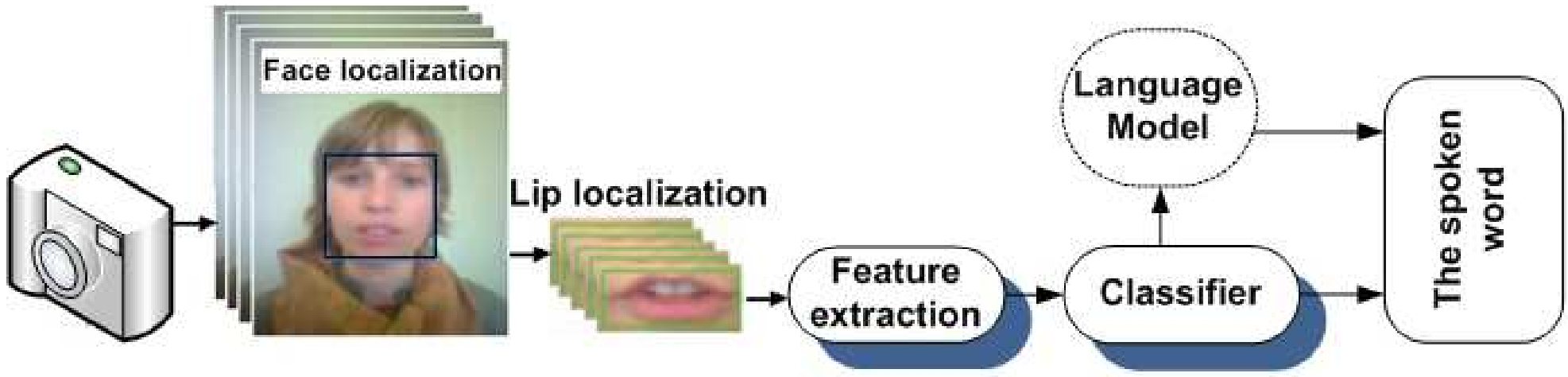

Fig. 1. A typical VSR system.

The last stage is the core of the system in which the visual features are extracted and the words are recognised. Unlike the visemic approach, this study proposes an holistic approach to tackle the VSR problem, where the system recognizes the whole word rather than just parts of it. In the proposed system, a word is represented by a signature that consists of several signals or feature vectors (feature matrix), e.g. height of the mouth, mutual information, etc.
Each signal is constructed by temporal measurements of its associated feature. The mouth height feature, for instance, is measured over the time period of a spoken word. This approach is referred to as the "visual words" (VW) approach. A language model is an optional step that can be used to enhance the performance of the system.

### 1.1 Human lip reading skills

Lip reading is not a contemporary invention; it was practised as early as 1500 AD, and probably before that time. The first successful lip reading teacher was the Spanish Benedictine monk, Pietro Ponce, who died in 1588. Lip reading teaching subsequently spread to other countries. The German Samuel Heinecke opened the first lip reading school in Leipzig in 1787. The first speech reading conference was held at Chautauqua, USA in 1894 (Bruhn, 1920).
Several different methods have been described in the literature for human lip reading such as the Muller-Walle, Kinzie, and the Jena methods. The Muller-Walle method focuses on the lip movement to produce a syllable as part of words, and the Kinzie method divides lip reading teaching into 3 teaching levels, depending on the difficulty (beginners, intermediate and advanced) (De Land, 1931). Although only 50% or less of speech can be seen, the reader must guesstimate those words that he/she has missed. This was the core of the Jena method: training the eye and exercising the mind (De Land, 1931). However, regardless of the variety of known lip reading methods, all methods still depend on the lip movement that can be seen by the lip reader.

Potamianos et al. (2001) described a human speech perception experiment. A small number of human listeners were presented with the audio once and the audio and video of 50 database sequences from an IBM ViaVoice database single speaker, with different bubble noises added each time. The participants were asked to transcribe what they heard and viewed.

Potamianos et al.'s (2001) experiment is not a pure lip reading experiment, as its aim was to measure the effect of the visual cues on the human speech perception, rather than the perception of the speech without the audio. The experiment showed that human speech perception increases by seeing the video and watching the visual cues. The word error rate was reduced by 20% when participants viewed the video, showing that the human audio-visual speech perception is about 62% word accuracy. According to the previous study, about 30% of the participants were non-native speakers, and this is one of the reasons why the recognition rate was very low, despite both the audio and video signals being revealed.

A human lip reading experiment was conducted in this study to roughly measure the human ability for lip reading, and the amount of information that can be seen from speech. Four video sequences from the PDA Database (Morris, et al., 2006) were used in this experiment, 2 males and 2 females; each video spoke 10 digits; the digits and their sequences are different from one video to another, and the audio signals were removed from the four videos. Fifty five participants were asked to transcript what each video spoke; each participant can play each video up to 3 times, so participants would have enough time to decide what the spoken digits were, and they would not be fooled by the speed of the video. These videos were uttering only digits, {1,2,3,…,9}, the participants were informed about this domain (the speech subject), hence it is much easier for humans to read lips if they know the subject of the talk, and also it mimics automatic lip reading experiments since the recognizer algorithm knows in advance and is trained on all the classes (the words) to be classified; the average word recognition rate for all the participants was 53%. See Table 1.

| Subject | Result |
|---|---|
| 1 (Female) | 61% |
| 2 (Male) | 50% |
| 3 (Male) | 37% |
| 4 (Female) | 63% |
| Average | 53% |

Table 1. Human lip reading results.

As can be seen from Table 1, some videos were easier to read than others (61% and 63% for videos 1 and 4 respectively), where some other videos have less information for lip readers, or those people by nature either speak faster than normal, or do not produce enough information for the lip readers. We can notice that the females give more information for the readers; it is, of course, difficult to substantiate such a claim because this is a small experiment using a small number of videos, so it is too early to draw such conclusions with such evidence. The most important thing that this experiment can reveal so far is the overall human lip reading ability, which is 53%. Another interesting thing to mention is that different people also have different abilities to perceive speech from visual cues only. In this experiment the best lip reader result was 73%, while the worst was 23%. These experiments illustrate the variation in individual lip reading skills, and the variation in individual ability to produce a clear readable visual signal, which would add to the challenge of designing an automatic lip reading system.

The human ability for lip reading varies from one person to another, and depends mainly on guessing to overcome the lack of visual information. Needless to say, lip readers need to have a good command of the spoken language, and in some cases the lip reader improvises and uses his/her knowledge of the language and context to pick the nearest word that he/she feels fits into the speech. Moreover, human lip readers benefit from visual cues detected outside the mouth area (e.g. gestures and facial expressions). The complexity and difficulties of modelling these processes present serious challenges to the task of automatic visual speech recognition.

### 1.2 In-house video database

Some of the methods discussed in this chapter are evaluated using an in-house video database (Hassanat, 2009). This database consists of 26 participants of different races and nationalities (Africans, Europeans, Asians and Middle Eastern volunteers). Each participant recorded 2 videos (sessions 1 and 2) at different times (about a 2 month period in time between the two recordings). The participants were 10 females and 16 males, distributed over different ethnic groups: 5 Africans, 3 Asians, 8 Europeans, and 10 Middle Eastern participants. Six of the males had both beard and moustache, and 3 males had moustache only.

The videos were recorded inside a normal room, which was lit by a 500-watt light source, using Sony HDR-SR10E high definition (HD) 40GB Hard Disc Drive Handy-cam Digital Camcorder – 4 Mega Pixels. The videos were de-interlaced then compressed using Intel IYUV codec, converted to AVI format, and resized to (320 x 240) pixels, because it is easier to deal with AVI format, and it is faster for training and analyzing the videos with smaller frame sizes. Each person in each recorded video utters non-contiguous 30 different words five times, which are numbers (from 0-9), short look-alike words (*knife, light, kit, night, fight*) and (*fold, sold, hold, bold, cold)*, long words: (*appreciate, university, determine, situation, practical*) and five security related words (*bomb, kill, run, gun, fire*).

### 1.3 Chapter overview

This chapter consists of 7 sections. In the first section, we presented a brief introduction to the VSR and human ability to read lips, and briefly described a typical VSR system. Section 2 briefly reviews automatic lip reading literature and describes some of the (state-of-the-art) approaches to VSR. Section 3 presents different approaches to face detection/localization. Lip localization approaches are reviewed in section 4. Section 5 is dedicated to the features extraction and recognition method. Some experimental results are presented in section 6. The chapter summary and some conclusions are discussed in section 7.

## 2. VSR literature review

Most of the work done on VSR came through the development of AVSR systems, as the visual signal completes the audio signal, and therefore enhances the performance of these systems. Little work has been done using the visual only signal. Most of the proposed lip reading solutions consist of two major steps, feature extraction, and Visual speech feature recognition. Existing approaches for feature extraction can be categorised as:

1. **Geometric features-based approaches** - obtain geometric information from the mouth region such as the mouth shape, height, width, and area.
2. **Appearance-based approaches** - these methods consider the pixel values of the mouth region, and they apply to both grey and coloured images. Normally some sort of dimensionality reduction of the region of interest (ROI) (the mouth area) is used such as the principal component analysis (PCA), which was used for the Eigenlips approach, where the first n coefficients of all likely lip configurations represented each Eigenlip.
3. **Image-transformed-based approaches** - these methods extract the visual features by transforming the mouth image to a space of features, using some transform technique, such as the discrete Fourier, discrete wavelet, and discrete cosine transforms (DCT). These transforms are important for dimensionality reduction and to redundant data elimination.
4. **Hybrid approaches**, which exploit features from more than one approach.

## 2.1 Geometric features-based approaches

A geometric features-based approach includes the first work on VSR done by Petajan in 1984, who designed a lip reading system to aid his speech recognition system. His method was based on using geometric features such as the mouth's height, width, area and perimeter (Petajan, 1984).

Another recent work in this category is the work done by (Werda et al., 2007), where they proposed an Automatic Lip Feature Extraction prototype (ALiFE), including lip localization, lip tracking, visual feature extraction and speech unit recognition. Their experiments yielded 72.73% accuracy of French vowels, uttered by multiple speakers (female and male) under natural conditions.

## 2.2 Appearance-based approaches

Eigenlips are the compact representation of mouth Region of Interest using PCA. This approach was inspired by the methods of (Turk & Pentland, 1991), and first proposed by (Bregler & Konig, 1994). Another Eigenlips-based system was investigated by (Arsic & Thiran, 2006), who aimed to exploit the complementarity of audio and visual sources. (Belongie & Weber, 1995) introduced a lip reading method using optical flow and a novel gradient-based filtering technique for the features extraction process of the vertical lip motion and the mouth elongation respectively.

In a more recent study, (Hazen et al., 2004) developed a speaker-independent audio-visual speech recognition (AVSR) system using a segment-based modelling strategy. This AVSR system includes information collected from visual measurements of the speaker's lip region using a novel audio-visual integration mechanism, which they call a segment-constrained Hidden Markov Model (HMM). (Gurban & Thiran, 2005) developed a hybrid SVM-HMM system for audio-visual speech recognition, the lips being manually detected. The pixels of down-sampled images of size 20 x 15 are coupled to get the pixel-to-pixel difference between consecutive frames. (Saenko et al., 2005) proposed a feature-based model for pronunciation variation to visual speech recognition; the model uses dynamic Bayesian network DBN to represent the feature stream.

(Sagheer et al., 2006) introduced an appearance-based lip reading system, employing a novel approach for extracting and classifying visual features termed as "Hyper Column Model" (HCM). (Yau et al., 2006) described a voiceless speech recognition system that employs dynamic visual features to represent the facial movements. The system segments the facial movement from the image sequences using motion history image MHI (a spatio-temporal template). The system uses discrete stationary wavelet transform (SWT) and Zernike moments to extract rotation invariant features from MHI.

### 2.3 Image-transformed-based approaches

(Lucey & Sridharan's, 2008) work was designed to be posing invariant. Their audio-visual automatic speech recognition was designed to recognize speech regardless of the pose of the head, the method starting with face detection and head pose estimation. They used the pose estimation method described by (Viola & Jones, 2001). The pose estimation process determines the visual feature extraction to be applied either on the front face, the left or the right face profile. The visual feature extraction was based on the DCT, which was reduced by the linear discriminative analysis (LDA), and the feature vectors were classified using HMM.

A very recent study which also fits into this category was done by (Jun & Hua, 2009), where they used DCT for feature extraction from the mouth region, in order to extract the most discriminative feature vectors from the DCT coefficients. The dimensionality was reduced by using LDA. In addition, HMM was employed to recognize the words.

### 2.4 Hybrid approaches

(Neti et al., 2000) proposed an audio-visual speech recognition system, where visual features obtained from DCT and active appearance model (AAM) were projected onto a 41 dimensional feature space using the LDA. Linear interpolation was used to align visual features to audio features.

A comparative Viseme recognition study by (Leszczynski & Skarbek, 2005) compared 3 classification algorithms for visual mouth appearance (Visemes): 1) DFT + LDA, 2) MESH + LDA, 3) MESH + PCA. They used two feature extraction procedures: one was based on normalized triangle mesh (MESH), and the other was based on the Discrete Fourier Transform (DFT), the classifiers designed by PCA and LDA.

Yu (2008) made VSR the process of recognizing individual words based on a manifold representation instead of the traditional visemes representation. This is done by introducing a generic framework (called Visual Speech Units) to recognise words without resorting to Viseme classification.

The previous approaches can be further classified depending on their recognition and /or classification method. Researchers usually use dynamic time warping (DTW), e.g. the work done by Petajan. Artificial neural networks (ANN), e.g. the work done by Yau et al. and Werda et al.. Dynamic Bayesian Network (DBN), e.g. the work done by Belongie and Weber, and support vector machines (SVM), e.g. the work done by Gurban and Thiran, and Saenko et al.

The most widely used classifier in the VSR literature is the hidden Markov models (HMM). Methods that use HMM include Bregler and Konig; Neti, et al.; Potamianos et al.; Hazen et al.; Leszczynski and Skarbek; Arsic and Thiran; Sagheer, et al.; Lucey and Sridharan; Yu; and Jun and Hua.

Each of the previous approaches has its own strengths and weaknesses. Sometimes the data reduction methods cause the loss of a considerable amount of related data, while using all the available information takes a much longer processing time, and not necessarily to obtain better results due to video or image-dependent information. More effort should be invested to propose any combination of the different approaches, to trade the disadvantages of each individual approach.

Most of the previous studies on VSR contain promising solutions, especially when combining an audio signal with a video signal. Although most of these systems rely on a clean visual signal (Saenko et al., 2005), still, for visual alone speech reading systems or subsystems, they have a high word error rate (WER). Sometimes WER is more than 90% for large vocabulary systems (Hazen et al., 2004; Potamianos, et al., 2003) and a range of 55% to 90% for small vocabulary systems (Yau et al., 2006). The main reason behind this high WER is that VSR problems represent a very difficult task by nature, as the visual side provides little information about speech. Other reasons that increase WER include: Large variations in the way that people speak (Yau et al., 2006), errors in pre-process steps of VSR systems such as face detection and lips localization, visual appearance differences between individuals, particularly, in speaker-independent systems, the visemes problems, and other general problems like light conditions and video quality.

## 3. Face detection

Face detection is an essential pre-processing step in many face-related applications (e.g. face recognition, lips reading, age, gender, and race recognition). The accuracy rate of these applications depends on the reliability of the face detection step. In addition, face detection is an important research problem for its role as a challenging case of a more general problem, i.e. object detection.

The most common and straightforward example of this problem is the detection of a single face at a known scale and orientation. This is a nontrivial problem, and no method has yet been found that can solve this problem with 100% accuracy. Factors influencing the accuracy of face detection include variation in recording conditions/parameters such as pose, orientation, and lighting. However, there are several algorithms and methods that deal with this problem, attaining various accuracy rates under varied conditions. Most existing schemes are based on somewhat restrictive assumptions. Some of the most successful methods used 20×20 (or so) pixel observation window across the image for all possible locations, scales, and orientations. These methods include the use of support vectors machines (Osuna et al., 1997), neural network (Rowley et al., 1998) or the maximum likelihood approach based on histograms of feature outputs (Schneiderman and Kanade, 2000). Others use a cascaded support vector machine (Romdhani, et. al., 2004). Some researchers use the skin colour to detect the face in coloured images (Garcia, & Tziritas, 1999).

In their study, (Yang et. al., 2002) classified face detection methods in still images into four categories:

1. Knowledge-based methods. These methods require human knowledge about facial features.
2. Feature invariant approaches. Designed to find structural features that are not affected by the general problems as with the face detection process, such as pose and light conditions. The targeted features vary from one researcher to another, but mostly they concentrated on facial features, texture, skin colour, or a combination of the previous features.

3. Template matching methods. Using one or more patterns to describe a typical face, then comparing this pattern with the image to find the best correlation between the pattern and a window in the targeted image. These templates can be predefined templates or deformable templates.
4. Appearance-based methods. Like the previous approach, but the template is not previously declared, rather it is learned from a set of images, then the learned template is used for detection. A variety of methods fill in this gap, such as: Eigenface (e.g. Eigenvector decomposition and clustering), distribution-based (e.g. Gaussian distribution), see (Sung & Poggio, 1998; Samaria, 1994), Neural networks, support vector machines, Hidden Markov Model, Naïve Bayes classifier, and information-theoretical approach.

Others classified face detection methods into two approaches: features-based approaches and image-based approaches (Hjelmas, & Low, 2001). The problem with most of these methods is that they are very sensitive to variation in light conditions and complex backgrounds. However, the one proposed by Rowley et al. (1998) for face detection is one of the best face detection methods created so far, and is used as a benchmark test by many researchers.

Rowley's et al. (1998) method for face detection consists of two stages, first applying a neural network-based filter that receives a 20 x 20 pixel region of the input image, and output values ranging from -1 to 1, which means non-face or face respectively. Assuming that faces in an image are upright and looking at the camera, the filter is applied to all locations in the image, to obtain all the possible locations of the face. To solve the scale problem, the input image is repeatedly sub-sampled by a factor of 1.2. The input image is pre-processed before inputting the proposed system. Light correction and a histogram equalizer were used to equalize the intensity values in each window.

The second stage focuses on merging overlapping detections and the arbitration process. The same face is detected many times with adjacent locations, the centre of these locations being considered as the centre of the detected face, and if two face locations are overlapped, the one with the highest score is considered the face location. Multiple networks were used to improve detection accuracy by ANDing (or ORing) the output of two networks over different scales and positions. Rowley's system was evaluated using 130 images containing 507 faces, the images having been collected from newspaper pictures, photographs and the World Wide Web. To train the system on false examples, 1000 images with random pixel intensities were generated. The detection rate of this system ranged from 78.9% - 90.5% depending on the arbitration used (ANDing or ORing).

Rowley's scheme is tested on our video database and detected al l the faces in the videos, and therefore was used for the purpose of this study.

## 4. Lip detection

Over the last few decades, the number of applications that are concerned with the automatic processing/analysis of human faces has grown remarkably. Many of these applications have a particular interest in the lips and mouth area. For such applications a robust and real-time lips detection/localization method is a major factor contributing to their reliability and success. Since lips are the most deformable part of the face, detecting them is a nontrivial problem, adding to the long list of factors that adversely affect the performance of image processing/analysis schemes, such as variations in lighting conditions, pose, head rotation, facial expressions and scaling.

The lips and mouth region are the visual parts of the human speech production system; these parts hold the most visual speech information, therefore it is imperative for any VSR system to detect/localize such regions to capture the related visual information, i.e. we cannot read lips without seeing them first. Therefore, lip localization is an essential process for any VSR system.

### 4.1 Existing trends in lip detection

Many techniques for lips detection/localization in digital images have been reported in the literature, and can be categorized into two main types of solutions:

1. *Model-based* lips detection methods. Such models include "Snakes", Active Shape Models (ASM), Active Appearance Models (AAM), and deformable templates.
2. *Image-based* lips detection methods. These include the use of spatial information, pixel colour and intensity, lines, corners, edges, and motion.

#### 4.1.1 Model-based lip detection methods

This approach depends on building lip model(s), with or without using training face images and subsequently using the defined model to search for the lips in any freshly input image. The best fit to the model, with respect to some prescribed criteria, is declared to be the location of the detected lips. For more about these methods see (Cootes & Taylor, 1992; Kass, et. al., 1987; Yuille, et. al, 1989).

#### 4.1.2 Image-based lip detection methods

Since there is a difference between the colour of lips and the colour of the face region around the lips, detecting lips using colour information attracted researchers' interest recently because of simplicity, not being time consuming, and the use of fewer resources, e.g. low memory, allowing many promising methods for lip detection using colour information to emerge. The most important information that researchers focus on include the red and the green colours in the RGB colour system, the hue of the HSV colour system, and the component of the red and blue in the YCbCr colour system. Some researchers used more information from the lip edges and lip motion. A well known mixed approach is called the "hybrid edge" (Eveno, et. al., 2002).

### 4.2 The adopted lip detection method

Among the model-based lip-detection methods, the active shape models are the most common and best performing technique for lip detection. However, this approach is basically affected by factors such as facial hair and does not meet some of the functional requirements (e.g. in terms of speed). "The implementation of Active Shape Model ASM was always difficult to run in Real-Time" (Guitarte, et. al., 2003). The same thing applies to the AAM approach. In fact, AAM is slower than ASM.

In grey level images and under diffuse lighting conditions, the external border of the lip is not sharp enough (see Figure 2a), and this makes the use of techniques based on the information provided by these images alone ineffective (Coianiz, et. al., 1996). Moreover, the ASM and the AAM algorithms are sensitive to the initialization process. When initialization is far from the target object, they can converge to local minima (see Figure 2a).

Other problems, like the appearance of the moustache, beard, and accessories, contribute to the problems and challenges that model-based lip detection methods undergo. Figure 2b illustrates some of these problems.

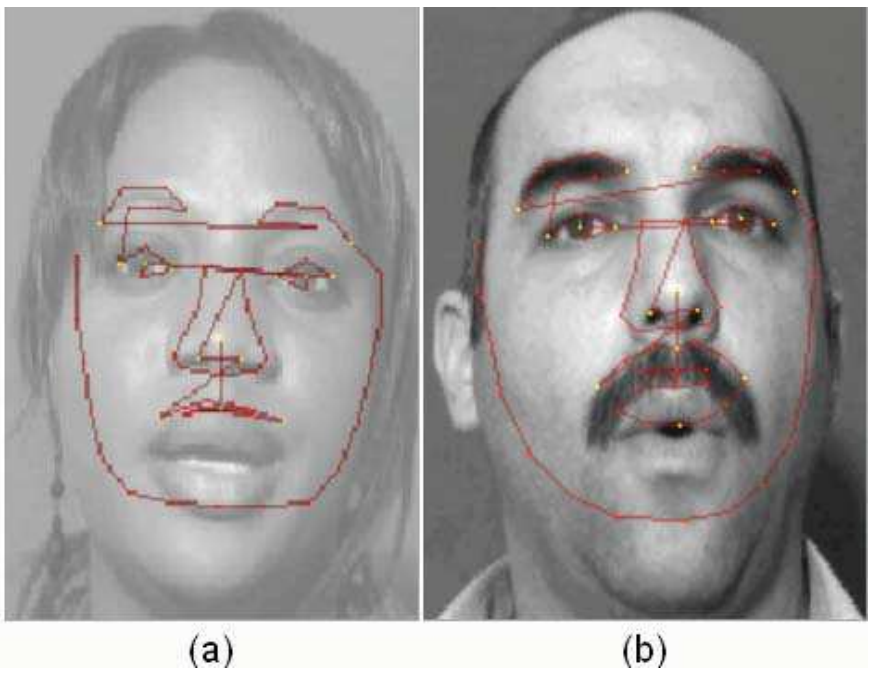

Fig. 2. Facial features detection using MASM*, (a) lip detection converges to local minima, (b) the effect of facial hair on ASM convergence.

Since the VSR problem needs several pre-processing steps, e.g. face and lips detection, it is vital for the VSR system to have faster solutions for these steps in order that the final solution can work in real time.

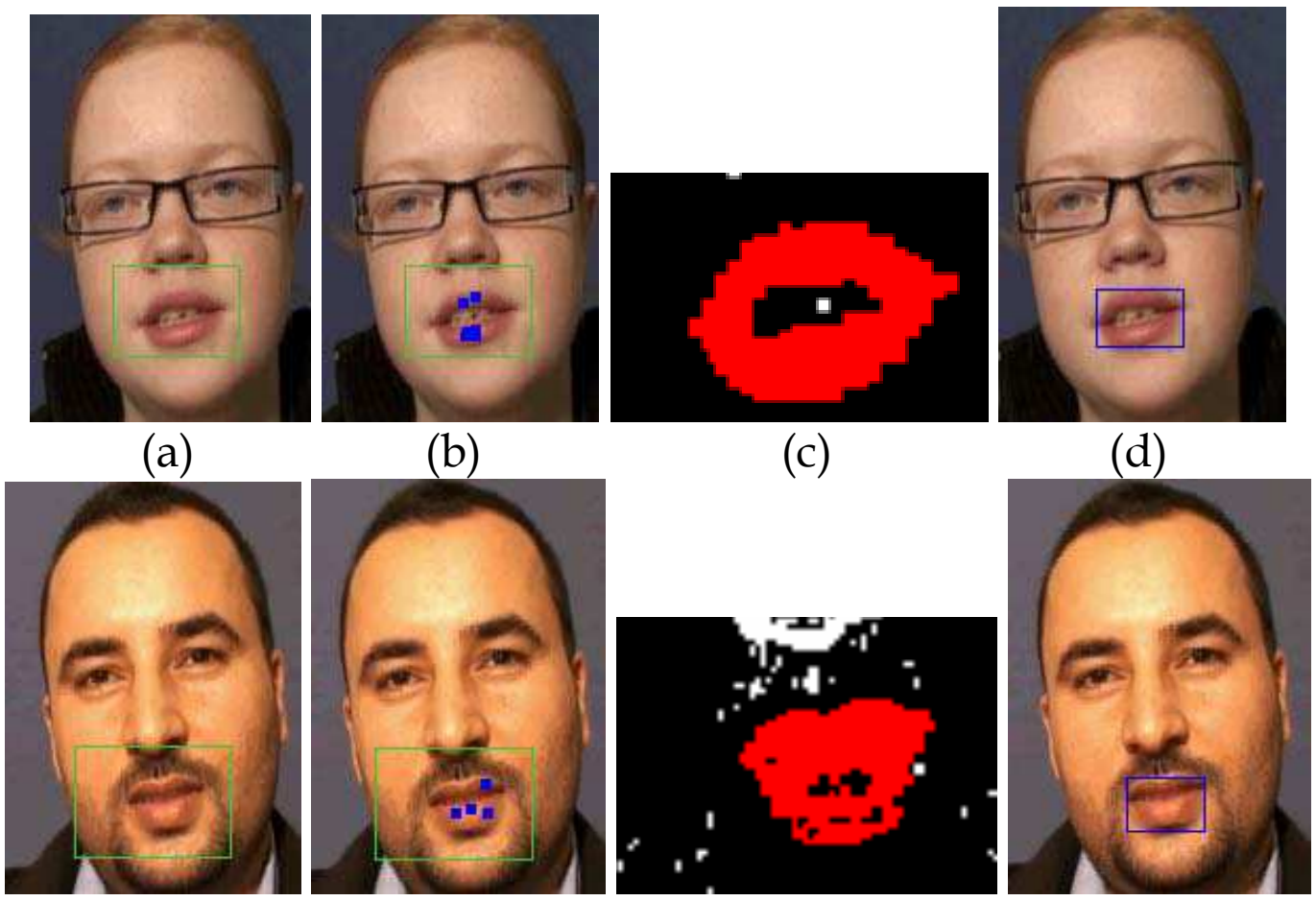

Fig. 3. The different stages of the "nearest colour", a) face detection followed by ROI defining, b) initial clustering using the YCbCr, c) binary image of ROI resulting from the nearest colour algorithm, d) final lip detecting.

* "MSAM" is a state-of-art ASM free library, developed by (Milborrow & Nicolls, 2008), and can be downloaded from the following link: http://www.milbo.users.sonic.net/stasm/download.html

In order to overcome the above-mentioned difficulties, we shall use a colour-based method (for lip localization in our study), in spite of being vulnerable to variations in light conditions. Such an effective method is described in our previous work (Hasanat, 2009). This method is based on using the YCbCr approach to find at least any part of the lip as an initial step. Then we use all the available information about the segmented lip-pixels such as r, g, b, warped hue, etc. to segment the rest of the lip. The mean is calculated for each value, then for each pixel in ROI, and Euclidian distance from the mean vector is calculated. Pixels with smaller distances are further clustered as lip pixels. Thus, the rest of the pixels in ROI will be clustered (to lip/non-lip pixel) depending on their distances from the mean vector of the initial segmented lip region. See Figure 3.

The method was evaluated on 780,000 frames of the in-house database; the experiments show that the method localizes the lips efficiently, with high level of accuracy (91.15%).

## 5. Features extraction and recognition

VSR systems require the analysis of feature vectors, which are extracted from the speech-related visual signals, in ROI in the sequence of the speaker face frames while uttering the spoken word/speech. Ideally, the required feature representations of words must capture specific visual information that is closely associated with the spoken word, to enable the recognition of the word and distinguish it from other words. Unlike the visemic approach, the visual words technique depends on finding a signature for the whole word, instead of recognizing each part (Viseme) of the word alone. To find such a signature, or a signal for each word, we need to find a proper way of extracting the most relevant features, which play an important role in recognizing that word.

An appearance-based approach to visual speech feature extraction ignores the fact that mouth appearance varies from one person to another (even when two persons speak the same word). Thus, using the appearance-based feature extraction alone does not take individual differences into consideration, and leads to inaccurate results. Moreover, appearance-based feature extraction methods mostly lack robustness in certain illumination and lighting conditions (Jun & Hua, 2009).

In this chapter, we adopt the hybrid-based approach, and we expand on the list of features beyond traditionally adopted ones such as the height and width of the speaker's lips. Indeed, there is valuable information encapsulated within the ROI that has a significant association with the spoken word, e.g. the appearance of the tongue and teeth in the image during the speech. The appearance of the teeth (for instance) occurs while uttering specific phonemes (the dentals and labio–dentals). At the same time, focusing only on the image-based features (appearance and transformed-based features) yields image-specific features, and it is sometimes difficult to generalize about those features on other videos or speakers. These results are backed up by (Jun & Hua, 2009).

The visual signal associated with a phoneme is rather short and hence their visual features are extracted from "representative" image frames. However, the visual signals associated with words are of longer duration involving tens of frames that vary in many ways. Hence the need to supplement/modify the set of features used in a visemic system by including some features relating to variation of frames along the temporal axis. There are many ways to represent such features, but we shall include two seemingly obvious features: an image quality parameter that measures the deviation/distortion of any frame from its predecessor, as well as the amount of mutual information between a frame and its predecessor. Such features are expected to compensate for the fact that many words do share some phonemes.

The list of features adopted in this chapter is by no means exclusive, but it was limited out of a desire to minimize the number for efficiency purposes and to a manageable set of features for which their impact on the accuracy of the intended VSR system can be estimated experimentally. The following is the proposed list of features that will be extracted from the sequences of the ROIs of the mouth areas during the uttering of the word (see Figure 4):

1. The height (H) and width (W) of the mouth, i.e. ROI height and width (geometric-based features).
2. The mutual information (M) between consecutive frames ROI in the discrete wavelet transform (DWT) domain (image-transformed-based features based on temporal information).
3. The image quality value (Q) of the current ROI with reference to its predecessor measured in the DWT domain (image-transformed-based features based on temporal information).
4. The ratio of vertical to horizontal features (R) taken from DWT of ROI (image-transformed-based features based on temporal information).
5. The ratio of vertical edges to horizontal edges (ER) of ROI (image-transformed-based features).
6. The amount of red colour (RC) in ROI as an indicator of the appearance of the tongue (image-appearance-based features).
7. The amount of visible teeth (T) in the ROI (image-appearance-based features).

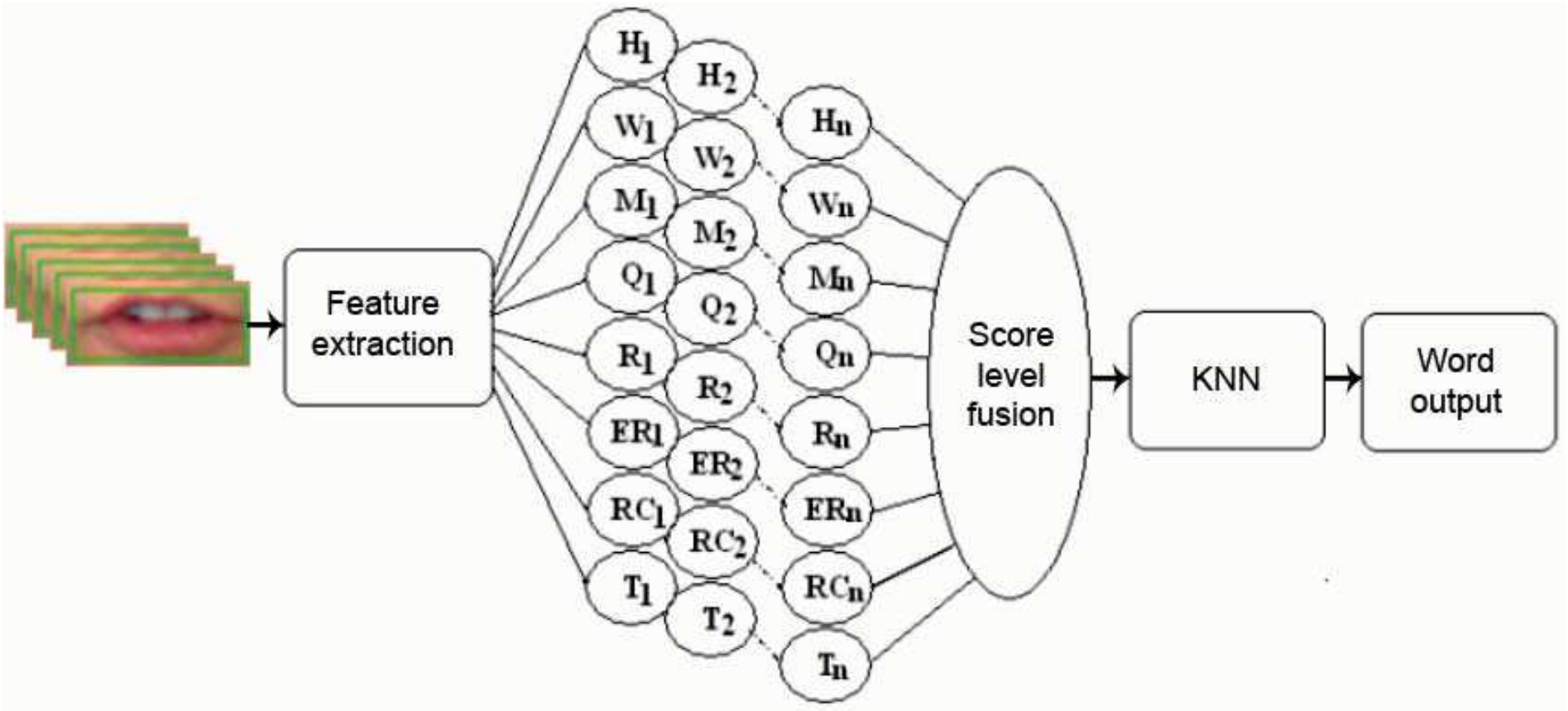

Fig. 4. The proposed feature extraction and recognition method.

As can be noticed from Figure 4, for each spoken word, eight feature vectors of length *n* (number of frames) are extracted, forming 8 different signals. This feature extraction method produces 8 signals for each uttered word, creating 8-dimensional feature space. Those signals maintain the dynamic of the spoken word, which contains a good portion of information; on the contrary, the visemic approach does not take into consideration the dynamic movement of the mouth and lips to produce a spoken word (Yu, 2008).

Accordingly, for each word we would extract a time-series of 8-dimensional vectors. The main difficulties in analysing of these time-series stem from the fact that their lengths not only differ between the spoken words, but also differ between different speakers uttering the same word and between the different occasions when the same word is uttered by the same speaker. In what follows we describe each of the 8 features. We assume that for each frame of a video, the speaker's face and lips are first localized (see sections 3 and 4) to determine the ROI from which these features are extracted (see Figure 5).

### 5.1 The height and width features of the mouth

Some VSR studies used lip contour points as shape features to recognise speech. For example, Wang et al. (2004) used a parameter set of a 14 points ASM lip to describe the outer lip contour. In addition, Sugahara et al. (2004) employed a sampled active contour model (SACM) to extract lip shapes. Determining the exact lips contour is rather problematic due to the little differences in the pixel values between the face and the lips. Here, we argue that it is not necessary (redundant) to use all or some of the lip's contour points to define the outer shape of the lips, where the height and width of the mouth backed up with a bounding ellipse is enough to approximate the real outer contour of the lips (see Figure 5).

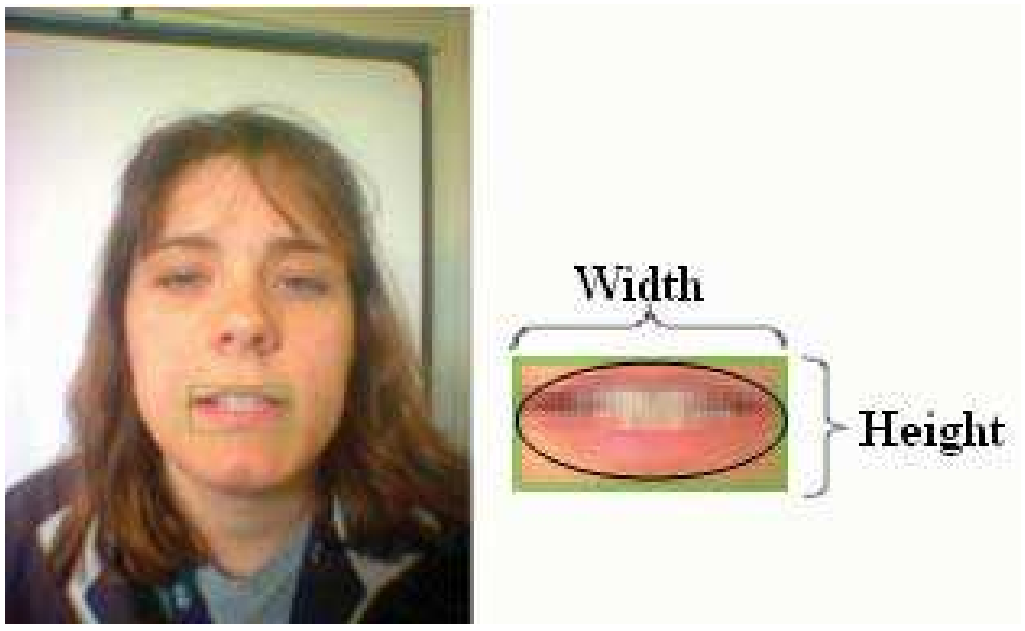

Fig. 5. Lips geometric feature extraction; width and height.

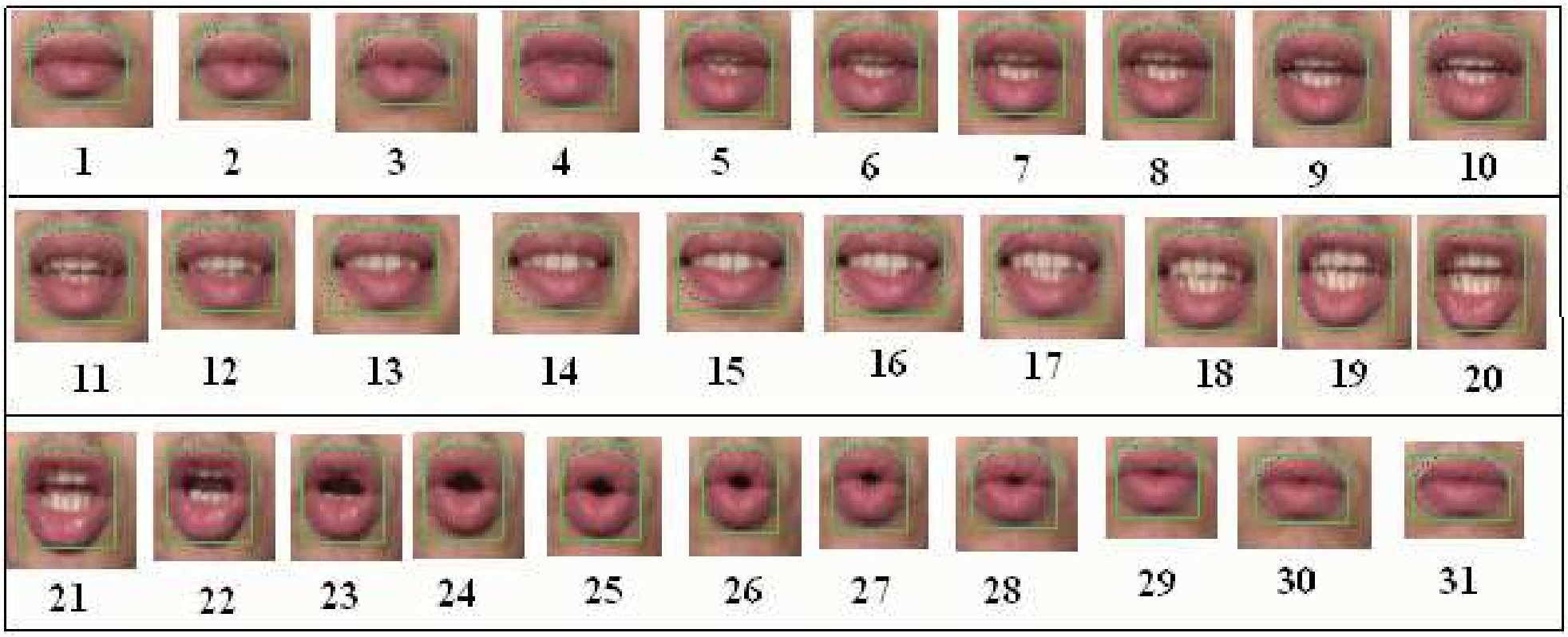

Fig. 6. The change of the mouth shape while uttering the word "Zero", the blue dotted ellipse shows the approximated lip contour using the ellipse assumption.

In our proposal, the height and width of the mouth area are determined by an approximation of the minimal rectangular box that contains the mouth area. The corners of the detected mouth box will be eliminated using the assumption that the mouth shape is the largest ellipse inside the minimal box. This assumption also helps in reducing redundant data when extracting the other features in the scheme. Sometimes lips are not horizontally symmetric due to different ways of speaking, the ellipse assumption forces such symmetries, and alleviates such differences between individuals (see Figure 6). These changes in height and width of the mouth create two signals (W and H) that represent changes during the time of uttering a specific word.

### 5.2 The mutual information (M) feature

Mutual information between 2 random variables X and Y defines the dependency of these variables, i.e. mutual information reveals how much X contains information about Y, and vice versa. Mutual information can be utilized to quantify the temporal correlation between frames of a video sequence, so it can calculate the amount of redundancy between any two frames. The temporal change of the appearance of ROI is caused by uttering a new/different phoneme. For example, the mouth appearance will change while switching from phoneme [ē] to phoneme [d] when uttering the word "feed". Therefore, it is sensible to use the mutual information to measure some aspects of the change in the mouth area between consecutive ROIs. The mutual information M between two random variables X and Y is defined by:

$$M(X;Y) = \sum_{x}\sum_{y} p(x,y)\log\left(\frac{p(x,y)}{p(x)p(y)}\right) \quad (1)$$

where p(x,y) is the joint probability mass function (PMF) of random variables X and Y (in our case mouth image (ROI) in current frame X, and previous mouth image in frame Y), p(x) and p(y) represent the marginal PMF of X and Y respectively. To use the mutual information formula, the size of both of the random variables must be the same, but because the height and width of ROI are changing over time while uttering different phonemes, consecutive ROIs might not be of the same size. To solve this problem, both ROIs are scaled to a predefined size, say 50 x 50 pixels.

Computing the mutual information in the spatial domain is inefficient and is influenced by many factors including the presence of noise and variation in lighting conditions. Instead, measuring the mutual information in the frequency domain provides a more informative mechanism to model changes between successive ROIs in different frequency sub-bands. Here we apply the DWT on both the current and the previous ROI. The mutual information formula is applied 4 times, one for each wavelet sub-band, and the average of the four values is taken as the mutual information feature for that frame or ROI (see Figure 7). Transforming both ROIs into the wavelet domain helps to reduce the effect of noise and variation in lighting conditions. For simplicity and efficiency, the DWT decomposition of the ROIs is implemented using the Haar filter.

### 5.3 The quality measure (Q) feature

There are many image quality measures proposed in the literature. Most of them attempt to find the amount of distortion in one image by referring to another image. Unlike the mutual information measurement, which attempts to measure the amount of dependency or similarity between two images, quality measures attempt to measure how different one image is from another.

Thinking again of the consecutive ROIs, a quality measure between them can tell something about change/distortion occurring due to an uttered phoneme. Therefore, any distortion in the current ROI, as compared to the previous ROI, is an indicator of changes in the structure of the mouth region. The amount of distortion can be measured by a quantitative quality measure, and considered as a feature at that frame or ROI.

This study utilizes a universal image quality index proposed by (Wang & Bovik , 2002) because it is a fast mathematical quality measure, and models image distortion as a combination of loss of correlation, luminance distortion, and contrast distortion. The quality measure Q is given by:

$$Q = \frac{4\sigma_{xy}\bar{x}\,\bar{y}}{(\sigma_x^2 + \sigma_y^2)[(\bar{x})^2 + (\bar{y})^2]} \tag{2}$$

where Q ∈ [-1,1],

$$\bar{x} = \frac{1}{N}\sum_{i=1}^{N} x_i \quad , \quad \bar{y} = \frac{1}{N}\sum_{i=1}^{N} y_i \quad , \quad \sigma_x^2 = \frac{1}{N-1}\sum_{i=1}^{N}(x_i - \bar{x})^2 \quad ,$$

$$\sigma_y^2 = \frac{1}{N-1}\sum_{i=1}^{N}(y_i - \bar{y})^2 \quad , and \quad \sigma_{xy}^2 = \frac{1}{N-1}\sum_{i=1}^{N}(x_i - \bar{x})(y_i - \bar{y})$$

The best value for Q is when there is no distortion in the current ROI compared to the previous ROI; the value then is equal to 1 or -1 and the maximum distortion is measured when Q = 0.

This formula is very sensitive to luminance, because it models image distortion as luminance distortion, as well as loss of correlation and contrast distortion. This problem is solved by using the same approach that was used for the mutual information feature, i.e. using the DWT decomposed ROIs. Again, 4 quality measures (Q) are computed, one for each wavelet sub-band (the HH, HL, LH, and the LL). Then the average of the four values is taken as the quality measure feature for that frame or ROI (see Figure 7). For compatibility, we also use the Haar filter and then both ROIs are scaled to 50 x 50 pixels. The average of both the mutual M, and the quality Q features is defined by:

$$M_i = \frac{M(LL_i;\mathrm{LL}_{i\ 1}) + M(HL_i;HL_{i\ 1}) + M(LH_i;LH_{i\ 1}) + M(HH_i;HH_{i\ 1})}{4} \tag{3}$$

$$Q_i = \frac{Q(LL_i;LL_{i\ 1}) + Q(HL_i;HL_{i\ 1}) + Q(LH_i;LH_{i\ 1}) + Q(HH_i;HH_{i\ 1})}{4} \tag{4}$$

where $M_i$ and $Q_i$ are the mutual and quality features at frame *i* respectively, $LL_i$, $HL_i$, $LH_i$ and $HH_i$ are the wavelet sub-bands of the current ROI, and $LL_{i-1}$, $HL_{i-1}$, $LH_{i-1}$ and $HH_{i-1}$ are the wavelet sub-bands of the previous ROI.

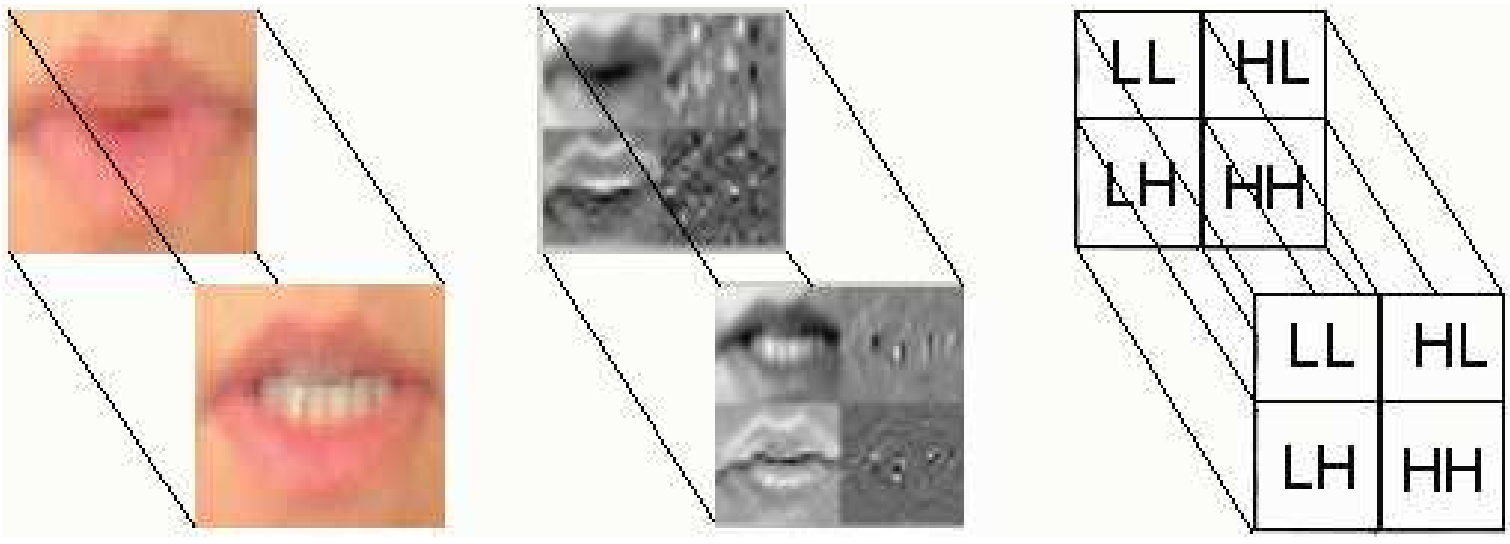

Fig. 7. (1st row) The previous mouth and its Haar wavelet, (2nd row) Current mouth and its Haar wavelet.

### 5.4 The ratio of vertical to horizontal features (R)

The DWT of an image I, using any wavelet filter, then histogram of the approximation sub-band LL approximates that of the original image while the coefficients in each of the three other sub-bands have a Laplacian distribution with 0 means (Al-Jawad, 2009). Moreover, in each non-LL-sub-band the further away from the mean a coefficient is, the more likely it is associated with a significant image feature such as edges/corners.

Here we adopt the above approach to identify feature-related pixels as the significant coefficients in the Non-LL sub-bands, i.e. the feature points are the ones with values greater than (median + standard deviation), and less than (median – standard deviation). The ratio (*R*) of the vertical features obtained from wavelet sub-band HL to the number of the horizontal ones gained from the LH is given by:

$$R = \frac{V}{H} \tag{5}$$

where V = number of vertical features, and H = number of horizontal features. Accordingly, by substituting V and H in equation 5, we get equation 6.

$$R = \frac{\sum_x \sum_y \begin{cases} 1 & (HL_{median} + \sigma_{HL}) \le HL(x,y) \le (HL_{median} + \sigma_{HL}) \\ 0 & otherwise \end{cases}}{\sum_x \sum_y \begin{cases} 1 & (LH_{median} + \sigma_{LH}) \le LH(x,y) \le (LH_{median} + \sigma_{HL}) \\ 0 & otherwise \end{cases}} \tag{6}$$

where $HL_{median}$ and $LH_{median}$ are the medians of the wavelet sub-band *HL* and *LH* respectively, *HL(x,y)* and *LH(x,y)* the intensity value at location (x,y) in both *HL* and *LH* wavelet sub-bands, $\sigma_{HL}$ and $\sigma_{LH}$ are the standard deviation in both of the mentioned sub-bands. Figure 8 demonstrates the correlation between the mouth appearance and its ratio (*R)* property while speaking.

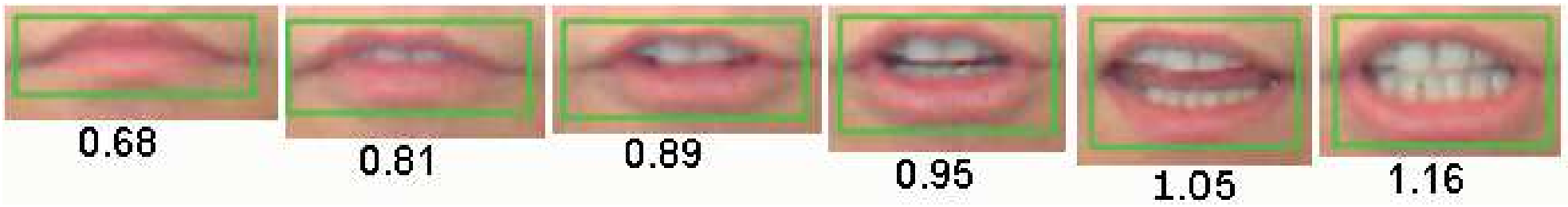

Fig. 8. The co-relation between the mouth appearance and its ratio (R).

As can be seen from Figure 8, the ratio R is high when there are a lot of vertical features of ROI compared to the horizontal ones (the mouths on the right), and R is low when vertical features are low and/or horizontal features are high (the mouths on the left).

### 5.5 The ratio of vertical edges to horizontal edges (ER)

The ratio of vertical edges to horizontal edges (ER) of ROI is obtained by using the Sobel edge detector. The summation of the absolute values of the vertical filter demonstrates the amount of vertical edges in the ROI. In addition, the summation of the absolute values of the horizontal filter demonstrates the amount of horizontal edges in the ROI. The ratio of the vertical edges to the horizontal ones is given by:

$$ER = \frac{\sum_{x=1}^{W}\sum_{y=1}^{H}\sum_{i=1}^{1}\sum_{j=1}^{1}\left|ROI(x+i,y+j)(S_v(i+1,j+1))\right|}{\sum_{x=1}^{W}\sum_{y=1}^{H}\sum_{i=1}^{1}\sum_{j=1}^{1}\left|ROI(x+i,y+j)(S_h(i+1,j+1))\right|} \quad (7)$$

where ROI(x,y) is the intensity value at the location (x,y) of the mouth region, W is the width of ROI, *H* is the height of ROI. *Sv*, and *Sh* are Sobel vertical and horizontal filters respectively.

When the mouth is stretched horizontally, the amount of horizontal edges increases, so ER decreases. When the mouth is opened, the amount of vertical edges tends to increase, and this increases the ER. Therefore, ER reveals something about the appearance of the mouth at a particular time.

### 5.6 The appearance of the tongue (RC)

Some phonemes like [*th*] involve the appearance of the tongue, i.e. moving the tongue and showing it helps to utter such phonemes. Therefore detecting the tongue in the ROI reveals something about the uttered phoneme and, by implication, the visual word.

However, it is difficult to model the tongue; the only available cue is its red colour. Therefore, the amount of red colour (RC) in the ROI will be taken to represent the appearance of the tongue, as well as the lip colour. Since the lip is captured within the ROI, the change of the red colour amount is then a cue for the appearance of the tongue. The different size of the tongue and lip from person to person is not problematic, hence all the features are scaled to the range [0,1], and the ratio of the red colour to the size of ROI is considered. This ratio can be calculated using the following equation:

$$RC = \frac{\sum_{x1}^{W}\sum_{y=1}^{H} red(ROI(x,y))}{(W)(H)} \quad (8)$$

where *red*(*ROI*(x,y)) is the red component value of the RGB colour system at the (x,y) of the mouth region, W is the width of ROI, H is the height of ROI.

### 5.7 The appearance of the teeth (T)

Some phonemes like [s] incorporate the appearance of the teeth, i.e. showing teeth helps to utter such phonemes. Therefore detecting teeth in the ROI is a visual cue for uttering such phonemes and enriches the visual words signatures. The major characteristic that distinguishes teeth from other parts of the ROI is the low saturation and high intensity values (Goecke, 2000). By converting the pixels values of ROI to 1976 CIELAB colour space (L*, a*, b*) and 1976 CIELUV colour space (L*, u*, v*), the teeth pixel has a lower a* and u* value than other lip pixels (Liew et. al., 2003). A teeth pixel can be defined by:

$$t = \begin{cases} 1 & a^* \leq (\mu_a - \sigma_a) \\ 1 & u^* \leq (\mu_u - \sigma_u) \\ 0 & otherwise \end{cases} \tag{9}$$

where $\mu_a, \sigma_a$ *and* $\mu_u, \sigma_u$ are the mean and standard deviation of a* and u* in ROI respectively. The appearance of the teeth can be defined by the number of teeth pixels in ROI. Therefore, the amount of teeth in ROI is given by:

$$T = \sum_{x=1}^{W} \sum_{y=1}^{H} t(x,y) \tag{10}$$

### 5.8 The classification process

All the previous features are normalized to the range [0,1] to alleviate the individual differences, and different scales of mouth caused by different distances from the camera, i.e. the different sizes of ROIs. For each property, a feature vector (a signal) is obtained to represent the spoken word from that feature perspective. Consequently, for each spoken word we get a feature matrix. The feature matrix has a fixed number of columns, but with a different number of rows, depending on the uttered word, and on the different speed of uttering words.

To compare signals with different lengths, we use the Dynamic Time Warping (DTW) method, and linear interpolation. For the fusion of the aforementioned features, we used score level fusion, which includes the use of each feature vector alone, using an empirical weighting technique to give different weights for the features, to capture the reliability of each feature vector, depending on how informative they are.

For each signal, the distances are measured with other signals from the training data, using DTW or Euclidian distance after linear interpolation, to overcome the different signal lengths. According to the K-Nearest-neighbour (KNN), the minimum k weighted averages are considered to predict the class (word) by announcing the maximum occurrence class in the nearest k as the predicted class.

## 6. Some experimental results

We evaluate the discussed VSR system using our in-house video database, in addition to the following main types of experiments:

1. **Speaker-dependent experiment:** this was conducted on each subject alone, all the test examples, and the training examples pertaining to the same subject (person). The main goal of this experiment is to test the way of speaking unique to each person, and each one's ability to produce a visual signal that was easily read. These experiments use leave-one example-out cross-validation protocol, test samples came from session 1 and training samples from session 2 for the same subject.
2. **Speaker-independent experiment:** In this type of experiment, the computer tests each subject against the rest of the subjects. Each time, one subject is taken out of the training set and is tested against the remaining subjects in the database. The training set does not contain any examples belonging to the tested subject. So the leave-one-subject-out cross-validation is used to evaluate the system. This type of experiment neglects the individual differences in appearance, and in the way of speaking.

| Subject | Speaker dependent | Speaker independent | Subject | Speaker dependent | Speaker independent |
|---|---|---|---|---|---|
| Female01 | **69%** | 27% | Male04 | **63%** | 16% |
| Female02 | **97%** | 49% | Male05 | **85%** | 36% |
| Female03 | **87%** | 35% | Male06 | **65%** | 19% |
| Female04 | **81%** | 39% | Male07 | **84%** | 36% |
| Female05 | **83%** | 26% | Male08 | **75%** | 27% |
| Female06 | **75%** | 43% | Male09 | **92%** | 41% |
| Female07 | **82%** | 31% | Male10 | **88%** | 31% |
| Female08 | **85%** | 29% | Male11 | **84%** | 53% |
| Female09 | **81%** | 41% | Male12 | **69%** | 26% |
| Female10 | **88%** | 42% | Male13 | **53%** | 15% |
| Male01 | **79%** | 43% | Male14 | **69%** | 33% |
| Male02 | **61%** | 15% | Male15 | **83%** | 39% |
| Male03 | **44%** | 23% | Male16 | **62%** | 28% |
| | **All** | | | **76.38%** | 33% |
| | Females | | | 83% | 36% |
| | Males | | | 72% | 30% |
| | Excluding moustache & beard | | | 77% | 33% |
| | Moustache & beard | | | 71% | 30% |

Table 2. VSR system Word recognition rates

It can be noticed from Table (2) that the overall WER of speaker-dependent experiments was (76.38%), and it was only 33% for the speaker-independent experiment. Our experiments show that the speaker-dependent word recognition rate is much higher than that of the speaker-independent; this claim is backed up by several researchers such as (Jun & Hua, 2009). Individual differences in the mouth appearance, and in the way of talking, produce different visual and audio signals for the same spoken word, which emphasizes that the visual speech recognition problem is a **speaker-dependent problem**.

We can notice also the negative effect of the facial hair on the results, when excluding subjects with moustaches and beards performance increased by 6% (from 71% to 77%). This explains the female's best results (83%). Moreover, the training set contains native and non-native speaker subjects, and each of the non-native speakers has his/her own way of uttering English words, for example the word "determine" is pronounced in 3 different ways by the non-native subjects, "di-tur-min", "de-teir-main" and "de-ter-men". This gives the training set different signatures for the same word, which confuses the recognition algorithm and contributes to the "bad examples" pool*. Moreover, the training set contains different ethnic groups, African, Asian, Middle Eastern and European; these groups are different in appearance. Furthermore, there are also differences in the appearance of males versus females, and the differences between age groups, i.e. different colours and shapes of the lips and mouth region. This variety leads to different features being extracted from the same word, which again contributes to the "bad examples" pool and leads to unexpected results.

Another interesting observation is the large difference between the individual ability to produce the visual signal while talking (word recognition rate varies from 44% to 97%). We found that some participants have less ability to produce this signal, i.e. they talk with minimum lip movement, which makes it a difficult task to read their lips, even by human intelligence. We termed those persons "visual-speechless persons" (VSP). In our experiments, we found that Male02 and Male13 are VSP (see Figure 9).

| Subject | Visual representation |
|---|---|
| Male02 | 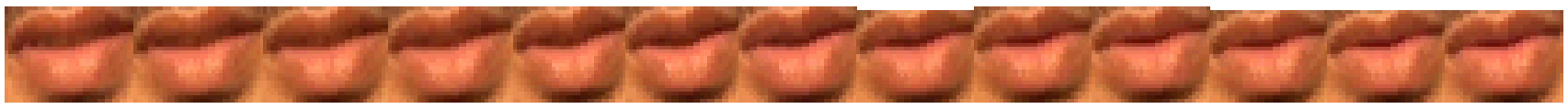 |
| Male13 | 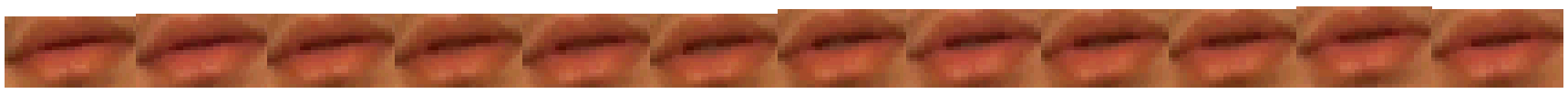 |
| Female2 | 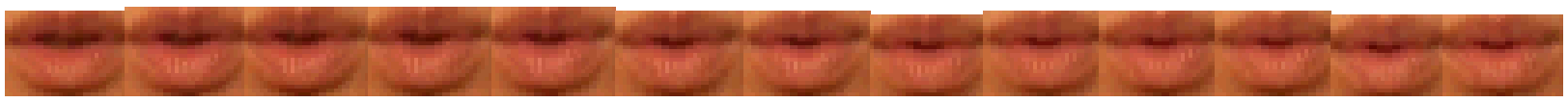 |

Fig. 9. Illustrating VSP concept, $1^{st}$ and $2^{nd}$ rows show the appearance of the word "two" uttered by two VSPs and the 3rd row shows the same word uttered by a normal person.

The previous Figure (9) shows that the VSPs do not produce clear visual signals. i.e. the appearance, shape and dynamic of ROI from the $1^{st}$ to the last frame, seems to be the same (unchanged to some extent), which makes it difficult to produce a unique signature for their visual speech, resulting in low WER for such subjects.

## 7. Chapter summary and conclusion

In this chapter, we described a complete VSR system, which includes face and lip detection/localization, features extraction and recognition. We evaluated the described scheme using two types of experiments, speaker-dependent and speaker-independent.

* Some examples in the training set, which are meant to represent some words, are closer to other words, e.g. outliers.

These experiments were carried out using a special database, which was designed for evaluation purposes. There were high word recognition rates for some subjects, and low ones for some others. Several reasons were found that affected the various results such as the appearance of facial hair, and the individual's aptitude to produce a clear visual signal. Some subjects produce weak signals (termed as VSP).
Results of the speaker dependents experiments were much better than that of the speaker independent. Therefore, we consider the VSR as speaker dependent problem, and to confirm such a result we need to further investigate VSR using different databases, and try to find some appearance invariant features, to minimize the effect of the visual appearance differences between individuals.
The major challenge for VSR is the lack of information in the visual domain, compared to the audio domain, perhaps because humans have yet to evolve to have need of a more sophisticated communication system. For example, it was sufficient for man's survival to use sound to warn friends if there was an enemy or a predator around without having to see them. Therefore, humans did not worry about producing a sufficient visual signal while talking. This major challenge, along with some others, opens the door for more research in the future, to compensate for the lack of information.

## 8. Acknowledgment

The author would like to acknowledge the financial support of Mutah university/Jordan (www.Mutah.edu.jo). In addition to professor Sabah Jassim for all his viable advices and discussions.

## 9. References

Al-Jawad, N., (2009). *Exploiting Statistical Properties of Wavelet Coefficients for Image/ Video Processing and Analysis Tasks*, PhD thesis, University of Buckingham, UK.

Arsic, I., & Thiran, J. (2006). Mutual Information Eigenlips For Audio-Visual Speech Recognition, *Proceedings of the 14th European Signal Processing Conference (EUSIPCO).*

Belongie, S., & Weber, M. (1995). Recognising Spoken Words from Lip Movement, *Technical Report CNS/EE248*, California Institute of Technology, USA.

Bregler, C., & Konig, Y. (1994). Eigenlips for Robust Speech Recognition, *Proceedings of ICASSP94*, Vol. II, Adelaide, Australia, 669-672.

Bruhn, M. E. (1920). *The Muller Walle Method of Lip Reading for the Deaf*, press of Thos. P. Nichols & Son Co. Lynn, Ma. USA.

Coianiz, T., Torresani, L., & Caprile, B., (1996). 2D deformable models for visual speech analysis, in DG Stork & ME Hennecke (Eds.), *Speechreading by Humans and Machines*, pp. 391-398.

Cootes, T. F. & Taylor, C. J., (1992). Active Shape Models - Smart Snakes, *Proceedings of the British Machine Vision Conference*, Springer-Verlag, pp. 266-275.

De Land F. (1931). *The Story of Lip Reading*, The Volta Bureau, Washington D. C., USA.

Eveno, N., Caplier, A., & Coulon, P. Y. (2002). A Parametric Model for Realistic Lip Segmentation, *Proceedings of ICARCV 02*, IEEE Press 3, pp. 1426–1431.

Garcia, C., & Tziritas, G., (1999). Face Detection Using Quantized Skin Color Regions Merging and Wavelet Packet Analysis, *IEEE Transactions On Multimedia*, Vol.1, No. 3, pp. 264-277.

Goecke, R., Millar, J.B., Zelinsky, A., & Robert-Ribes, J. (2000) Automatic extraction of lip feature points, *Proceedings of the Australian Conference on Robotics and Automation*, pp. 31-36.

Guitarte, J., Lukas, K., & Frangi, A.F. (2003). Low Resource Lip Finding and Tracking Algorithm for Embedded Devices, *ISCA Tutorial and Research Workshop on Audio Visual Speech Processing*, pp. 111-116.

Gurban, M. & Thiran, J. (2005). Audio-Visual Speech Recognition With A Hybrid Svm-Hmm System, *Proceedings of the 13th European Signal Processing Conference (EUSIPCO).*

Hassanat, A. B.A, (2009). *Visual Words for Automatic Lip-Reading*, PhD thesis, University of Buckingham, UK.

Hazen, T. J., Saenko, K., La, C.H., & Glass, J., (2004) A segment-based audio-visual speech recognizer: Data collection, development and initial experiments, *Proceedings of the International Conference on Multimodal Interfaces*, pp. 235-242.

Hjelmas, E., & Low, B.K., (2001). Face Detection: A Survey, *Computer Vision and Image Understanding*, vol. 3, pp. 236-274.

Jun, H. & Hua, Z. (2009). Research on Visual Speech Feature Extraction, *Proceedings of the International Conference on Computer Engineering and Technology*, Volume 2, pp. 499 – 502.

Kass, M., Witkin, A., & Terzopoulos, D. (1987). Snakes: Active contour models, *International Journal of Computer Vision*, volume 1, pp. 321-33.

Leszczynski, M., & Skarbek, W. (2005). Viseme Recognition – A Comparative Study, *Proceedings of IEEE International Conference on Advanced Video and Signal-Based Surveillance*.

Liew, A.W.C., Leung, S.H., & Lau, W.H., (2003). Segmentation of Color Lip Images by Spatial Fuzzy Clustering, *IEEE Transactions on Fuzzy Systems*, vol.11, no.4, pp. 542-549.

Lucey, P., & Sridharan, S. (2008). A Visual Front-End for a Continuous Pose-Invariant Lip-reading System, *Proceedings of the 2nd International Conference on Signal Processing and Communication Systems*, 15-17 December 2008, Australia, Queensland, Gold Coast.

Morris C., Koreman j., Sellahewa h., Ehlers j., Jassim s., Allano L., & Garcia-salicetti s. (2006). The SecurePhone PDA Database, Experimental Protocol and Automatic Test Procedure for Multimodal User Authentication, *technical report, Secure-Phone (1EC IST-2002-506883) project*, Version 2.1.

Neti, C., Potamianos, G. & Luettin, J. (2000). Audio-visual speech recognition, *Final Workshop 2000 Report, Center for Language and Speech Processing*, The Johns Hopkins University, Baltimore, MD.

Osuna, E., Freund, R., & Girosi, F. (1997). Training support vector machines: An application to face detection. *Proceedings of CVPR*, pp. 130–136.

Petajan, E. (1984). *Automatic lipreading to enhance speech recognition*, Ph.D. Dissertation, University of Illinois at Urbana-Champaign, USA.

Potamianos G., Neti C., Iyengar G., & Helmuth E. (2001). Large-Vocabulary Audio-Visual Speech Recognition by Machines and Humans, *Proceedings of EUROSPEECH,* pp. 1027-1030, Aalborg, Denmark, 2001.

Potamianos, G., Neti, C., Gravier, G., Garg, A., & Senior, A. W. (2003). Recent Advances in the Automatic Recognition of Audio-Visual Speech, Invited, *Proceedings of the IEEE*, vol. 91, no. 9, pp. 1306-1326.

Romdhani, S., Torr, P., Lkopf, B., & Blake, A. (2004). Efficient Face Detection by a Cascaded Support Vector Machine Expansion, *Royal Society of London Proceedings Series A*, vol. 460, pp. 3283-3297.

Rowley, H., Baluja, S.,& Kanade, T. (1998). Neural network-based face detection, *PAMI 20*, pp. 23–38.

Saenko, K., Livescu, K., Glass, J. & Darrell, T. (2005). Production Domain Modeling Of Pronunciation For Visual Speech Recognition, *Proceedings of ICASSP*, Philadelphia.

Sagheer, A., Tsuruta, N., Taniguchi, R. I. & Maeda, S. (2006). Appearance feature extraction versus image transform-based approach for visual speech recognition, *International Journal of Computational Intelligence and Applications*, Vol. 6, pp. 101–122.

Samaria, F.S., (1994). *Face Recognition Using Hidden Markov Models*, PhD thesis, Univ. of Cambridge, UK.

Schneiderman, H., & Kanade, T. (2000). A statistical method for 3d object detection applied to face and cars, *Proceedings of CVPR*, pp. 746–751.

Sung, K-K., & Poggio, T., (1998). Example-Based Learning for View-Based Human Face Detection, *IEEE Trans. Pattern Analysis and Machine Intelligence*, vol. 20, pp. 39-51.

Turk, M., & Pentland, A. (1991). Eigenfaces for recognition, *Journal of Cognitive Neurosci*, vol. 3. no.1, pp. 71-86.

Viola, P. & Jones, M. (2001). Rapid object detection using a boosted cascade of simple features, *Proceedings of the International Conference on Computer Vision and Pattern Recognition – CVPR*, Kauai, HI, USA, pp. 511–518.

Wang, Z., & Bovik, AC. (2002). A universal image quality index, *IEEE Signal Process Lett. 9*, pp. 81–84.

Wang, S.L., Lou, W.H., Leung, S.H. & Yan, H. (2004). A real-time automatic lip-reading system, *Proceedings of the IEEE International Symposium on Circuits and Systems*, Vancouver BC, Canada, vol. 5, pp. 101-104.

Werda, S., Mahdi, W., & Ben-Hamadou, A. (2007). Lip Localization and Viseme Classification for Visual Speech Recognition, *International Journal of Computing & Information Sciences*, Vol.5, No.1.

Yang, M.H., Kriegman, D., & Ahuja, N. (2002). Detecting Faces in Images: A Survey, *IEEE Trans. Pattern Anal. Mach. Intell.*, Volume 24, pp. 34 – 58.

Yau, W., Kumar, D., & Arjunan, S. (2006). Voiceless Speech Recognition Using Dynamic Visual Speech Features, *HCSNet Workshop on the Use of Vision in Human-Computer Interaction, (VisHCI 2006)*, pp. 93-101.

Yu, D. (2008). *The Application of Manifold based Visual Speech Units for Visual Speech Recognition*, PhD thesis, Dublin City University, Dublin, Ireland.

Yuille, A., Cohen, D. S. & Hallinan, P. W., (1989). Feature extraction from faces using deformable templates, *Proceedings of the IEEE Comput. Soc. Conf. Comput. Vision and Pattern Recogn*, pp. 104-109.

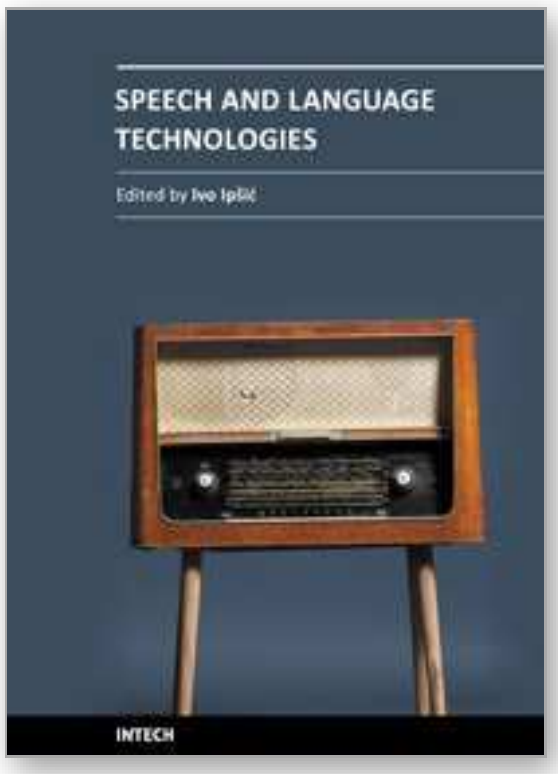

**Speech and Language Technologies**
Edited by Prof. Ivo Ipsic

This book addresses state-of-the-art systems and achievements in various topics in the research field of speech and language technologies. Book chapters are organized in different sections covering diverse problems, which have to be solved in speech recognition and language understanding systems. In the first section machine translation systems based on large parallel corpora using rule-based and statistical-based translation methods are presented. The third chapter presents work on real time two way speech-to-speech translation systems. In the second section two papers explore the use of speech technologies in language learning. The third section presents a work on language modeling used for speech recognition. The chapters in section Text-to-speech systems and emotional speech describe corpus-based speech synthesis and highlight the importance of speech prosody in speech recognition. In the fifth section the problem of speaker diarization is addressed. The last section presents various topics in speech technology applications like audio-visual speech recognition and lip reading systems.

**InTech Europe**
University Campus STeP Ri
Slavka Krautzeka 83/A
51000 Rijeka, Croatia
Phone: +385 (51) 770 447
Fax: +385 (51) 686 166
www.intechopen.com

**InTech China**
Unit 405, Office Block, Hotel Equatorial Shanghai
No.65, Yan An Road (West), Shanghai, 200040, China
中国上海市延安西路65号上海国际贵都大饭店办公楼405单元
Phone: +86-21-62489820
Fax: +86-21-62489821